\documentclass[conference]{ieeeconf}
\IEEEoverridecommandlockouts
\usepackage{cite}
\usepackage{amsmath,amssymb,amsfonts}
\usepackage{algorithmic}
\usepackage{graphicx}
\usepackage{textcomp}
\usepackage{xcolor}
\usepackage{soul} 
\usepackage{caption}
\usepackage{subcaption}
\usepackage[per-mode=symbol]{siunitx}
\usepackage{censor}
\usepackage{booktabs}
\usepackage{multirow}
\usepackage{flushend}
\usepackage{fancyhdr}

\fancypagestyle{customfirst}{
  \fancyhf{}
  \fancyhead[C]{\footnotesize \textit{\textcopyright 2025 IEEE.  Personal use of this material is permitted.  Permission from IEEE must be obtained for all other uses, in any current or future media, including reprinting/republishing this material for advertising or promotional purposes, creating new collective works, for resale or redistribution to servers or lists, or reuse of any copyrighted component of this work in other works.}\\Preprint version (Aug. 2025). This work has been accepted for publication in Humanoids 2025.}
  
}

\fancypagestyle{customrest}{
  \fancyhf{} 
  \fancyhead[C]{\footnotesize Preprint version (Aug. 2025). This work has been accepted for publication in Humanoids 2025.}
  
}
\def\censorcolor{white}
\let\svcensorrule\censorrule
\renewcommand\censorrule[1]{%
\textcolor{\censorcolor}{\svcensorrule{#1}}}

\def\BibTeX{{\rm B\kern-.05em{\sc i\kern-.025em b}\kern-.08em
    T\kern-.1667em\lower.7ex\hbox{E}\kern-.125emX}}

\StopCensoring 
\begin{document}

\title{\LARGE \bf
From Canada to Japan: How 10,000~km Affect User Perception in Robot Teleoperation
} 


\author{Siméon Capy$^{1}$, Thomas M. Kwok$^{2}$, Kevin Joseph$^{2}$, Yuichiro Kawasumi$^{3}$,\\
        Koichi Nagashima$^{3}$, Tomoya Sasaki$^{1}$,   
        Yue Hu$^{2}$ and Eiichi Yoshida$^{1}$
\thanks{$^{1}$Interactive Robotics Lab,
        Tokyo University of Science, 125-8585 Katsushika, Japan
        {\tt\small simeon.capy@rs.tus.ac.jp}}%
\thanks{$^{2}$Active and Interactive Robotics Lab,
        University of Waterloo, Waterloo, ON N2L 3G1, Canada
        }%
\thanks{$^{3}$Kawada Technologies, Inc.,
        111-0036 Taito, Japan
        }%
}


\maketitle

\thispagestyle{customfirst}  

\begin{abstract}
Robot teleoperation (RTo) has emerged as a viable alternative to local control, particularly when human intervention is still necessary. This research aims to study the distance effect on user perception in RTo, exploring the potential of teleoperated robots for older adult care. We propose an evaluation of non-expert users' perception of long-distance RTo, examining how their perception changes before and after interaction, as well as comparing it to that of locally operated robots. We have designed a specific protocol consisting of multiple questionnaires, along with a dedicated software architecture using the Robotics Operating System (ROS) and Unity. The results revealed no statistically significant differences between the local and remote robot conditions, suggesting that robots may be a viable alternative to traditional local control.
\end{abstract}


\section{Introduction}
Robot teleoperation (RTo) plays a crucial role in bridging the gap between current capabilities and the emergence of fully autonomous systems, particularly in scenarios where human intervention remains necessary. If global RTo experiments date back to the 1990s\cite{teleop97}, their practical and viable applications have emerged primarily since the early 21st century. A notable example is the first telesurgery performed between New York City (USA) and Strasbourg (France) in 2001\cite{mohan2021telesurgery-shortsurvey}. This progress has been further supported by advancements in mobile networks in recent years---where data transfer rates have increased from tens of kilobits per second with 2G to tens of gigabits per second with 5G\cite{5g}---teleoperated robots have become more viable and reliable. Indeed, latency remains a critical limitation for many tasks that cannot yet be performed autonomously\cite{kheddar-ge-jp}.

One domain where teleoperated robots could provide significant benefits is healthcare. Countries with ageing populations, such as Japan and Italy, are facing a rising demand for older adults care while experiencing a shortage of nursing personnel. According to WHO, by 2050, 22\% of the world population will be 60 years old or older\cite{oms}. This issue is further exacerbated by the increasing prevalence of medical deserts and the tendency of older adults individuals to remain in rural areas, where access to healthcare is limited. This is particularly true in vast country such as Canada where the distance between cities can be wide. Teleoperated robots could serve as a solution by assisting in patient care or supporting local medical staff through remote control from urban centres, where the majority of healthcare professionals are based; alleviating workforce shortage in the remote areas.

While the deployment of robots in retirement homes is not new---examples such as Paro and Pepper have been used as social companions\cite{paro,Tetsuya-Tanioka2019,Martinez-Martin2018}---the application of teleoperated robots in this context remains less common. Moreover, those examples are mostly for companionship and social interactions, whereas the teleoperated robots we are looking at in this paper have a primary focus on being functional and assistive. A noteworthy, albeit reversed, example is the Avatar Robot Café, where people with disabilities remotely operate OriHime robots to serve as waiters. This initiative enables patients to maintain social participation in society, even if they are unable to leave hospitals or their homes\cite{orylabAvatarRobot-url, orylabAvatarRobot-paper}. The use of teleoperated humanoid robots is also being explored as a solution in situations where robots need to replace humans without the need to design task-specific machines. Moreover, employing a humanoid robot in social interaction contexts can be more appropriate and may facilitate more natural interactions\cite{t-ro-survey}.

We can also mention other fields where teleoperated robots are widely used. For example, in hazardous environments, such as disaster response operations during the Fukushima nuclear incident\cite{tepcoTEPCOApplication} or the Notre-Dame de Paris fire\cite{ieeeParisFirefighters}. In addition to emergency applications, robot-assisted teleoperation has gained prominence in surgery since 2001\cite{surgery}. Another significant domain for RTo is the space industry, where environmental constraints such as isolation, vacuum, and radiation necessitate the use of teleoperated robots as a key solution, whether controlled from Earth or a spacecraft.

However, using remote robots can present challenges for the operator. It may alter task perception, particularly due to feedback latency or delays in command execution\cite{thomas-wearable}. Additionally, the overall experience is constrained by the system's technical limitations. Most teleoperation systems provide only visual and auditory feedback, often with a fixed or limited field of view\cite{thomas-gcn}. In applications requiring high precision, such as robotic surgery, the sense of touch can be simulated through haptic feedback, though it does not replicate the full richness of human tactile sensation.



This research aims to investigate user perceptions of remotely operated robots over long distances compared to locally controlled robots, with a focus on their application in the care of older adults. Our objective is to explore the potential of teleoperated robotic systems in mitigating workforce shortages in the healthcare sector. The contributions of this paper are twofold: (1) an evaluation of non-expert users’ perceptions of long-distance robot teleoperation, analysing changes before and after interaction and comparing them to local operation, and (2) the demonstration of a comprehensive user study protocol for teleoperation, incorporating multiple questionnaires and a custom software architecture built with ROS and Unity, adaptable to various robotic platforms.

\section{Related Work}
COVID-19 has presented an opportunity to advance the use of teleoperated robots in the healthcare industry. Yang et al. proposed a dual-armed robotic solution capable of delivering medication, performing basic auscultation, and interacting with medical equipment, with the goal of minimising direct contact between patients and medical staff\cite{covid19}. However, this line of research focuses primarily on technical implementation. Few studies have explored user perception of RTo. Sankra et al. conducted a survey on the perception of teleoperated robot interfaces (UI/UX), with nearly half of the systems studied were for general-purpose use, and 14\% dedicated to healthcare--—most of those in surgical contexts\cite{review-ui-ux}. Common evaluation tools used in these studies include NASA-TLX\cite{nasa-tlx} or the System Usability Scale (SUS)\cite{sus-brooke}. 

For studies specifically targeting user perception of teleoperated robot systems, Audonnet et al. evaluated user experience with UR3, UR5, and Baxter robots during locally performed pick-and-place tasks, using NASA-TLX, Negative Attitude Towards Robot Scale (NARS)\cite{nars}, and customised questionnaires to assess perception in a local teleoperation context with direct visual feedback\cite{audonnet2024breaking}. Moore et al. evaluated participants’ perception of robot size when operating the robots remotely versus with direct feedback. They also examined the influence of camera height in RTo. However, in both cases, the teleoperation was conducted within the same physical location\cite{rto-size-height}.

Another relevant study is from Wojtusch et al., who consulted a panel of experts to identify key human factors in a fictional delayed RTo scenario between the Earth and the Moon. A delay of 5 seconds was assumed, and the factors evaluated included situational awareness (Situation Awareness Rating Technique, SART\cite{sart}), workload (NASA-TLX), and user experience (User Experience Questionnaire, UEQ\cite{ueq}). Amongst their findings, they emphasised that the \textit{“reduction of unnecessary efforts \emph{[is]} crucial”} for the \textit{“success of the teleoperated task”}\cite{earth-moon}.

However, to the best of our knowledge, no prior studies have evaluated user perception of RTo over long distances.


\subsection*{Research questions}
The research questions (RQs) that guided this study were as follows:
\begin{itemize}
  \item[RQ1] \textit{Does distance influence users’ perception of teleoperated robots?}
  \item[RQ2] \textit{Does using teleoperated robots over long distances increase users’ perceived workload, or affect usability?}
\end{itemize}

To address these research questions, a system will be developed to enable long-distance teleoperation of the robots. In addition, a dedicated experimental protocol incorporating various evaluation tools will be designed.

\section{System Architecture}
\begin{figure}
    \centering
    \includegraphics[width=0.99\linewidth]{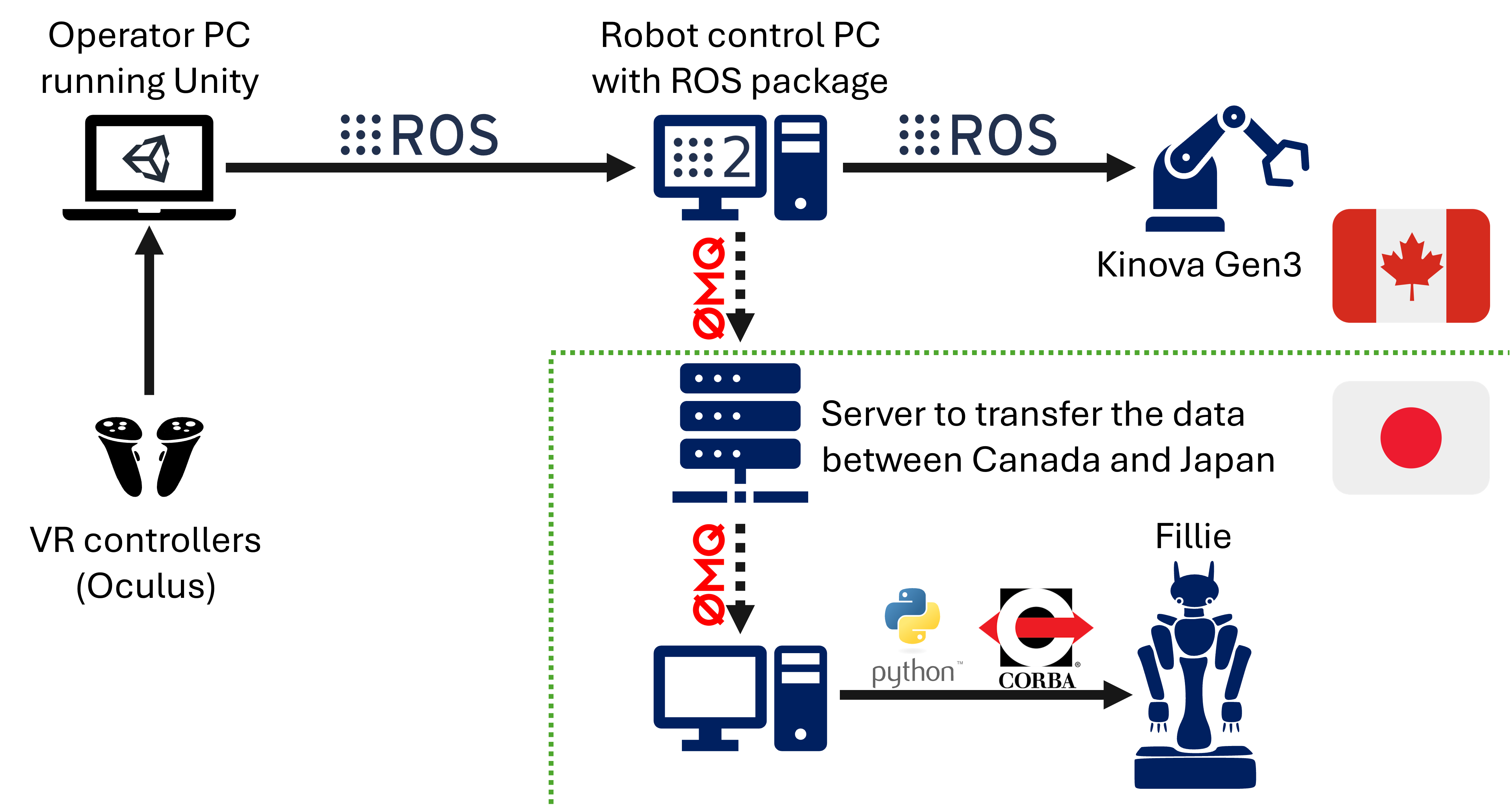}
    \caption{The green dotted line symbolises the separation between Japan and Canada. The solid arrows represent data communication within the local network (LAN), while the dotted arrows indicate communication over the Internet, with the protocol denoted by the corresponding logo.}
    \label{fig:project-architecture}
\end{figure}

\begin{figure}
     \centering
     \begin{subfigure}[b]{0.2\textwidth}
         \centering
         \includegraphics[width=\textwidth]{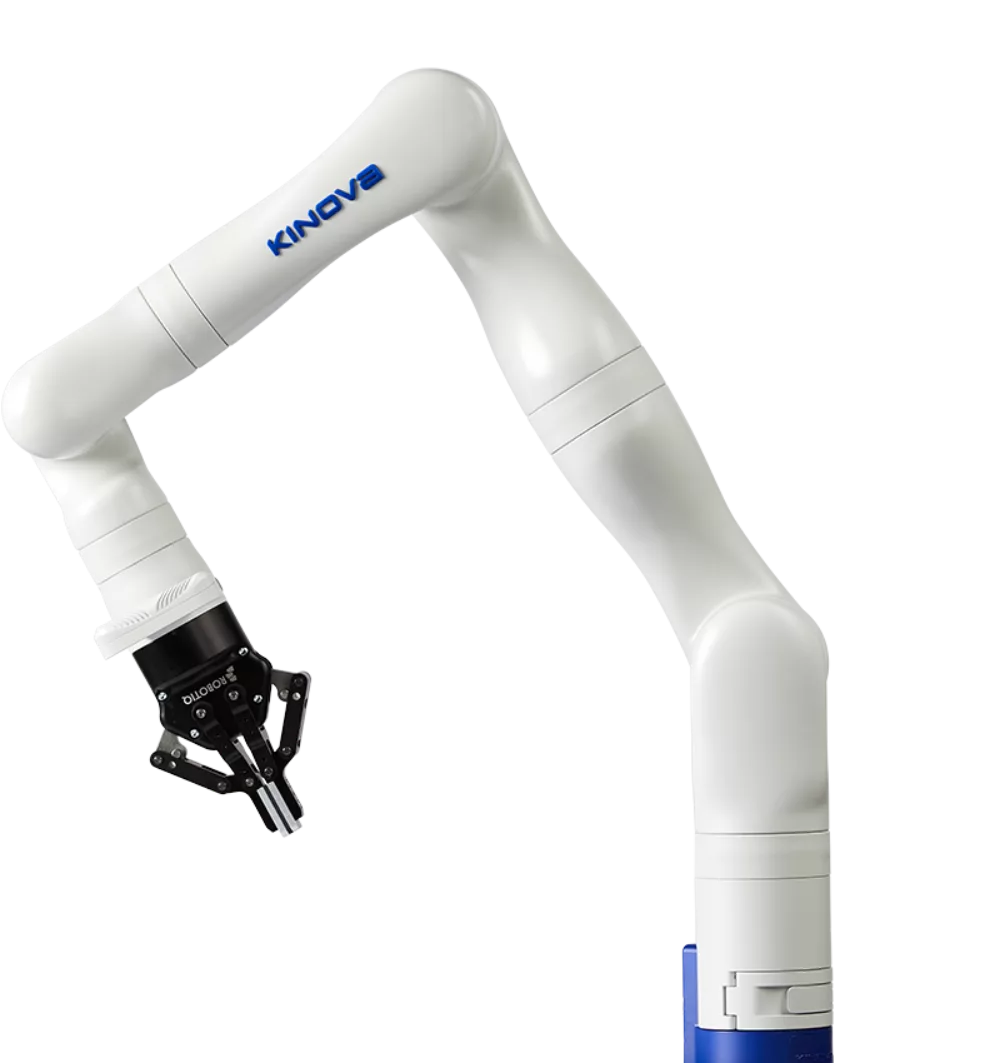}
         \caption{Kinova Gen3}
         \label{fig:kinova}
     \end{subfigure}
     \hfill
     \begin{subfigure}[b]{0.2\textwidth}
         \centering
         \includegraphics[width=\textwidth]{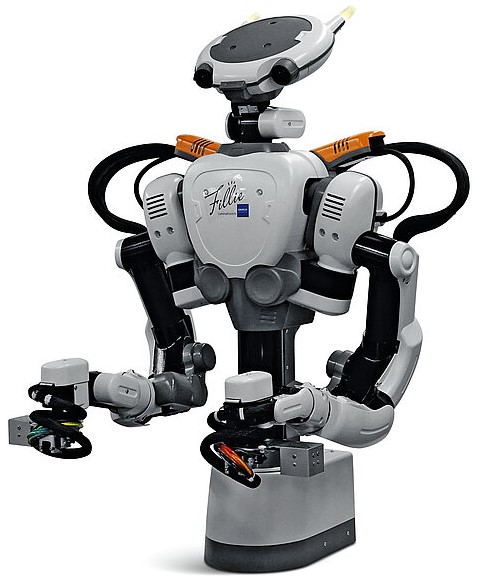}
         \caption{Fillie}
         \label{fig:fillie}
     \end{subfigure}
        \caption{The two robots used for the experiment, \ref{fig:kinova} was in Canada and \ref{fig:fillie} in Japan.}
        \label{fig:robots}
\end{figure}

Figure~\ref{fig:project-architecture} illustrates the project architecture, which consists of two main parts: the robot side and the operator side. These components communicate using either ROS~2 or ZMQ (ØMQ) protocols\cite{ros2,zmq}. Two different robots are employed: on the Canadian side, a Gen~3 from Kinova (called later Kinova) is used, while on the Japanese side, a NEXTAGE Fillie robot from Kawada Robotics (called later Fillie) is utilised (see Fig.~\ref{fig:robots}); due to logistical constraints, it was not possible to use the same robot for both locations. Although both robots were originally designed for industrial applications, and our aim is to use them for healthcare, Fillie has already demonstrated its applicability in social contexts\cite{capy2022expanding}. Each robot operates with a distinct communication protocol: Kinova interacts with the operator side via ROS, whereas Fillie utilises ZMQ, due to server constraints that supported only ZMQ. However, this does not impact the behaviour of the robots.

\subsection{Operator side}
The user controls the robot using VR controllers, specifically the Meta Quest~II in our setup, by transmitting the current position and orientation of the VR controller. The long-term goal of using our system with a VR headset led us to choose those controllers. This data is then sent to the robot, which uses it to determine the end effector (EE) position. The system is implemented in Unity, which communicates with the robot via ROS2, transmitting the following data: controller position, controller orientation (expressed as a quaternion for Kinova and as Euler angles for Fillie), gripper open/close commands, and a robot activation trigger. The latter is used to initiate robot movement, requiring the user to hold the trigger to keep the robot in motion. In this experiment, participants view a monitor (see Section \ref{sec:exp-protocol}); the VR headset is not used in the application developed for the experiment. However, it will be incorporated in the future to display the video stream.

\subsection{Robot side}
The robot-side system consists primarily of a local PC running our ROS2 package, developed in Python, which receives all messages from the operator side. For the local robot, the data are formatted appropriately and transmitted to the Kinova ROS package, which then controls the robot. In the case of Fillie, a ROS package is responsible for converting all ROS topics to ZMQ. These messages are then sent via the WebSocket protocol to a server in Japan, which simply relays them to the control computer in Japan. Finally, the commands are executed on the robot using Python and CORBA APIs.

\section{Experimental Protocol}
\label{sec:exp-protocol}
\begin{figure*}
     \centering
     \begin{subfigure}[b]{0.40\textwidth}
         \centering
         \includegraphics[height=\textwidth]{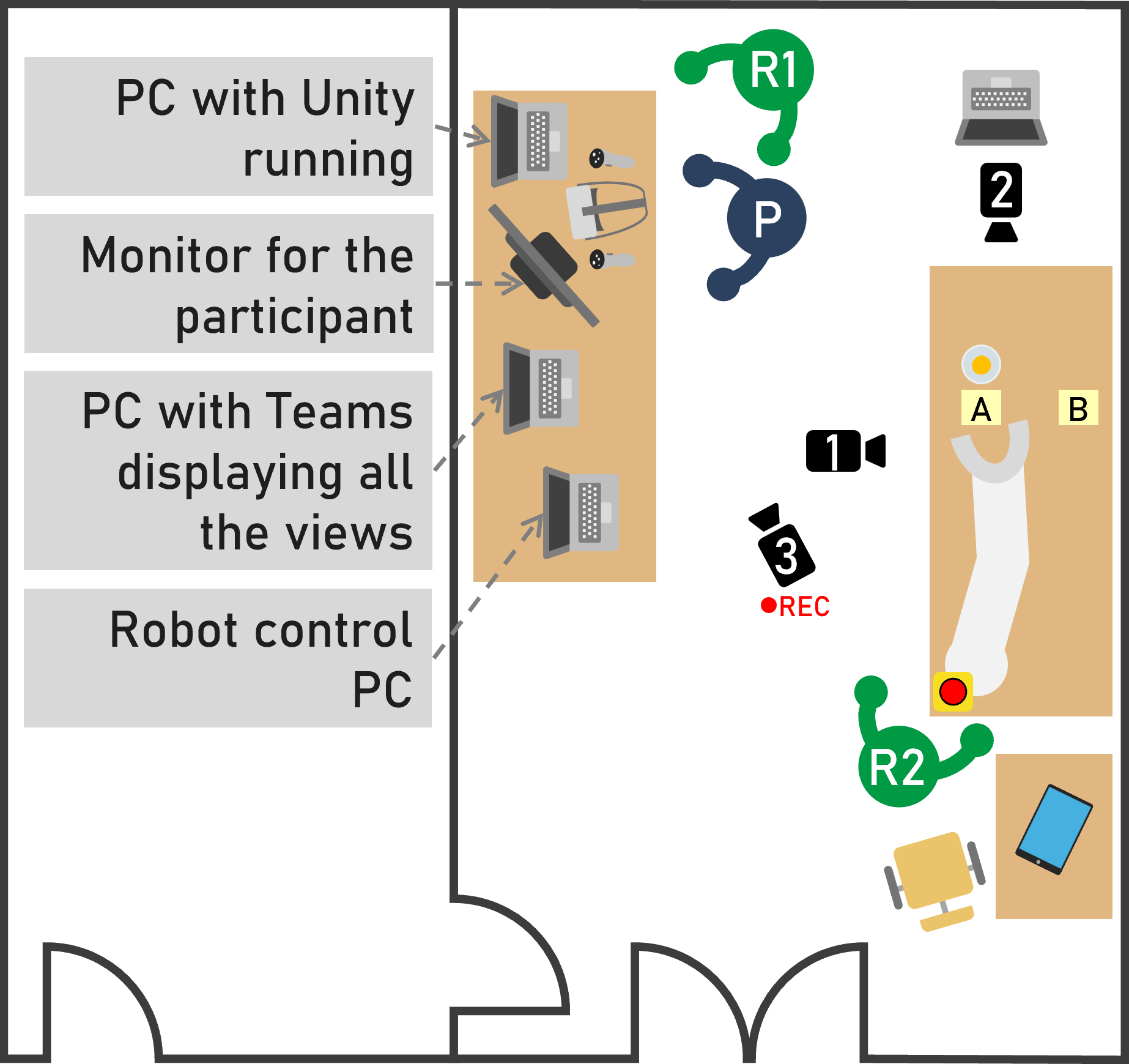}
         \caption{Experimental room in Canada (top view)}
         \label{fig:room-canada}
     \end{subfigure}
     \hfill
     \begin{subfigure}[b]{0.40\textwidth}
         \centering
         \includegraphics[height=\textwidth]{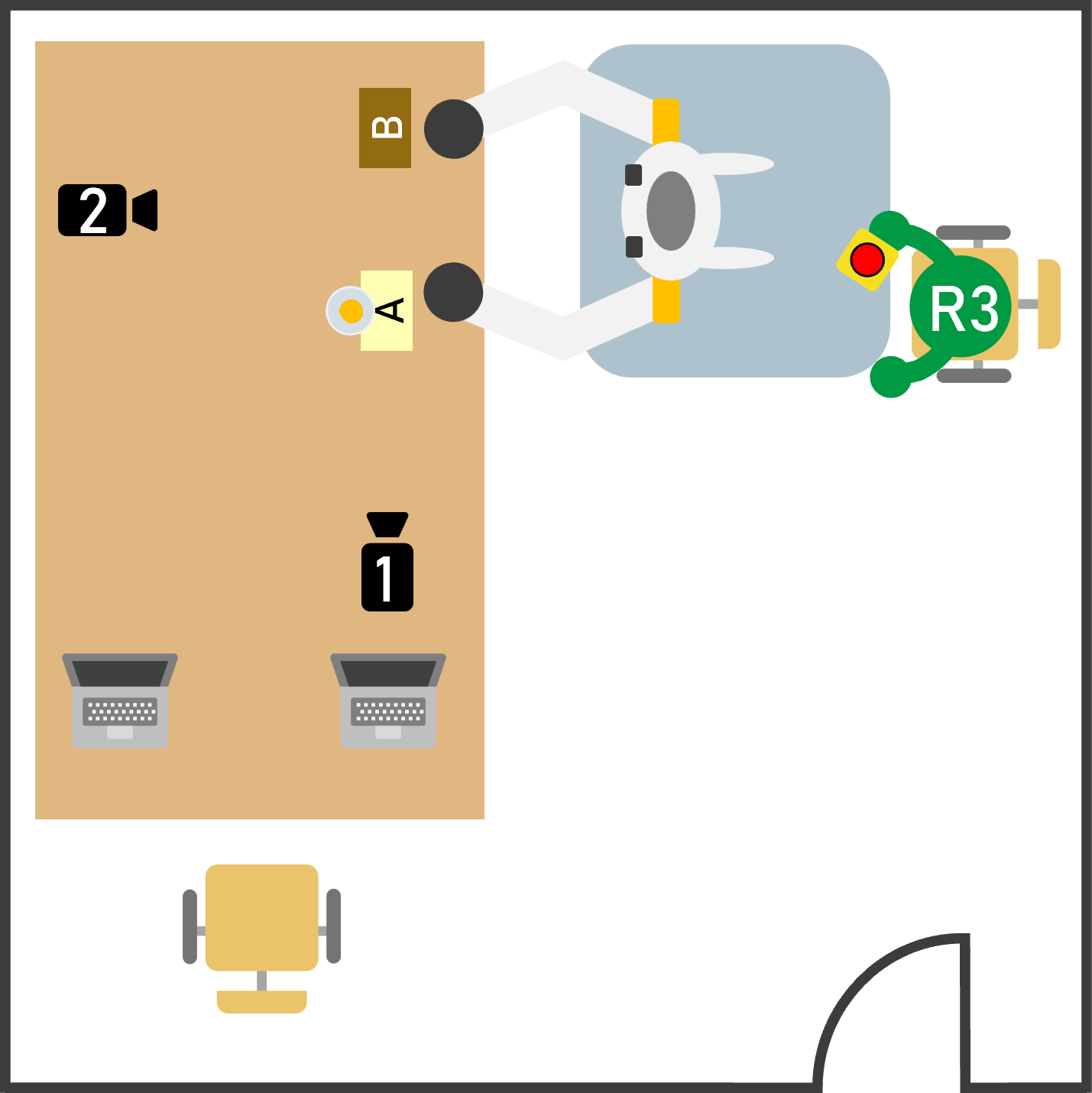}
         \caption{Experimental room in Japan (top view)}
         \label{fig:room-kawada}
     \end{subfigure}
        \caption{The participant (P) was located in Canada, facing a monitor displaying video feeds from cameras 1 and 2, while camera 3 recorded the participant. A researcher (R2 \& R3) was positioned near each robot (and visible on camera) to operate the emergency button, while another researcher (R1) was stationed close to the participant. Points A and B were approximately \SI{40}{\centi\meter} apart. The experiment was conducted using only Fillie’s right arm.}
        \label{fig:rooms}
\end{figure*}

The experiment took place in February and March 2025 at the University of Waterloo (UoW) in Canada and at Kawada Robotics in Japan. Participants were asked to control two robots, a 7-DoF robotic arm from Kinova, and a half-anthropomorphic robot from Kawada, see Fig.~\ref{fig:robots}. The control was done from Canada, and Kinova was located in the same room as the participants while Fillie was at Kawada offices, see Fig.~\ref{fig:rooms}. All participants provided informed consent before participation, and the experiment protocol was approved by the Research Ethics Committee of UoW under approval No.~46919 on 18 December 2024. Participants received a compensation of 15~CAD in cash for their participation.

Since our research focuses on the perception of robot teleoperation rather than task complexity, the specific task itself was not a primary variable. Therefore, we selected a standardised pick-and-place task involving a bottle. This task presents several advantages. First, it is relatively simple to perform, which helps prevent participant frustration due to excessive complexity. Second, it is a generic task that is easy to reproduce and adapt across different contexts. Finally, it is straightforward for researchers to observe and evaluate. Participants were required to move the bottle between two points separated by \SI{40}{\centi\meter}. Due to differences in the robots, the type of bottle and the grasping method varied: for the Kinova robot, a PET bottle was grasped from the side, whereas for the Fillie robot, a sauce bottle was grasped from the top, resembling a claw machine.

Participants were asked to complete a questionnaire at the beginning (Q0) of the experiment, after controlling the first robot (Q1), and at the end (Q2). This was done to track the evolution of perception. The choice of the different tools is based on the related works. Q0 is composed of demographic questions: age, gender, time since they are living in Canada, current job, highest diploma, and preferred hand; question about their experience regarding video games, robots and VR; and the Robotic Social Attributes Scale (RoSAS) questionnaire\cite{rosas}. The demographic questions were used to assess the representativeness of the participant sample and to identify any limitations. In addition, they support our aim to compare cultural aspects related to the perception of RTo, as outlined in the Future Work section (see Section~\ref{conclusion}). Q1 and Q2 are the same and contain the following tools: RoSAS, SUS, NASA-TLX (weighted as described in \cite{nasa-tlx-computation}) and four custom questions,
\begin{itemize}
    \item How would you describe your experience controlling the teleoperated robot? Please indicate whether it felt more like interacting with a real robot or similar to operating a video game or share any other impressions you had. (Likert scale + free text);
    \item The use of the system was… Responsive - Unresponsive (Likert scale);
    \item The use of the system was… Safe - Unsafe (Likert scale);
    \item I had the feeling I could harm the operator next to the robot unintentionally (Likert scale).
\end{itemize}
All the questions using a Likert scale used 7 values.

Due to the 14-hour time difference between Japan and Canada, the experiment commenced at 18:30 in Canada and 08:30 the following day in Japan, with \SI{30}{\minute} for each participant. The protocol was as follows: participants first read and signed the consent form, then completed questionnaire Q0. They then proceeded to the computer, where they received instructions before controlling one of the robots. The order of robot operation was alternated for each participant, to avoid bias order, with the first of each day always starting with the local robot to allow the Japanese partner time to prepare. Upon completion of the task with the first robot, they completed questionnaire Q1, after which they operated the second robot and subsequently answered questionnaire Q2.
\begin{figure}
    \centering
    \includegraphics[width=0.99\linewidth]{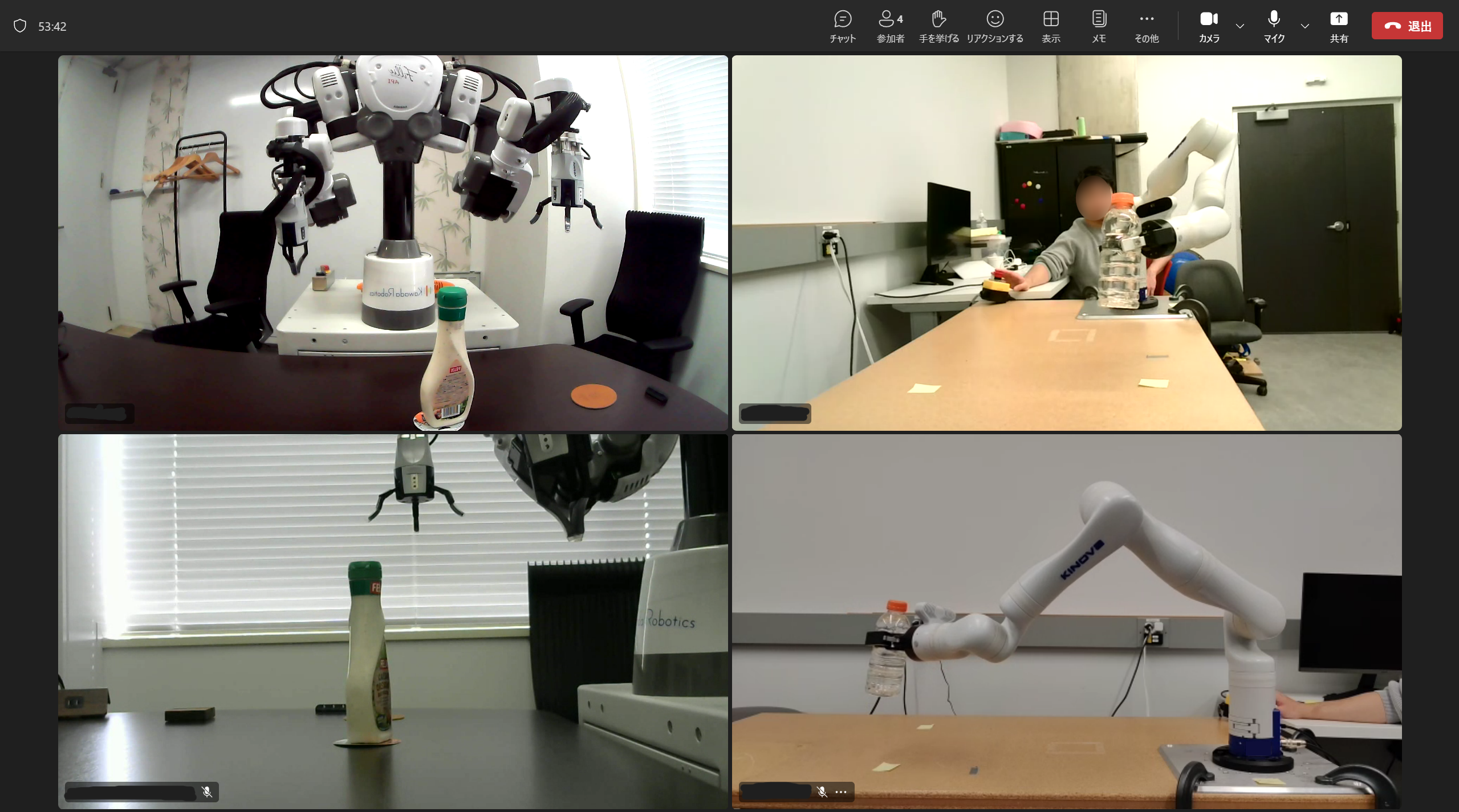}
    \caption{Control screen with the four views, two for each robot}
    \label{fig:teams-screen}
\end{figure}

\begin{figure*}
     \centering
     \begin{subfigure}[b]{0.49\textwidth}
         \centering
         \includegraphics[width=\textwidth]{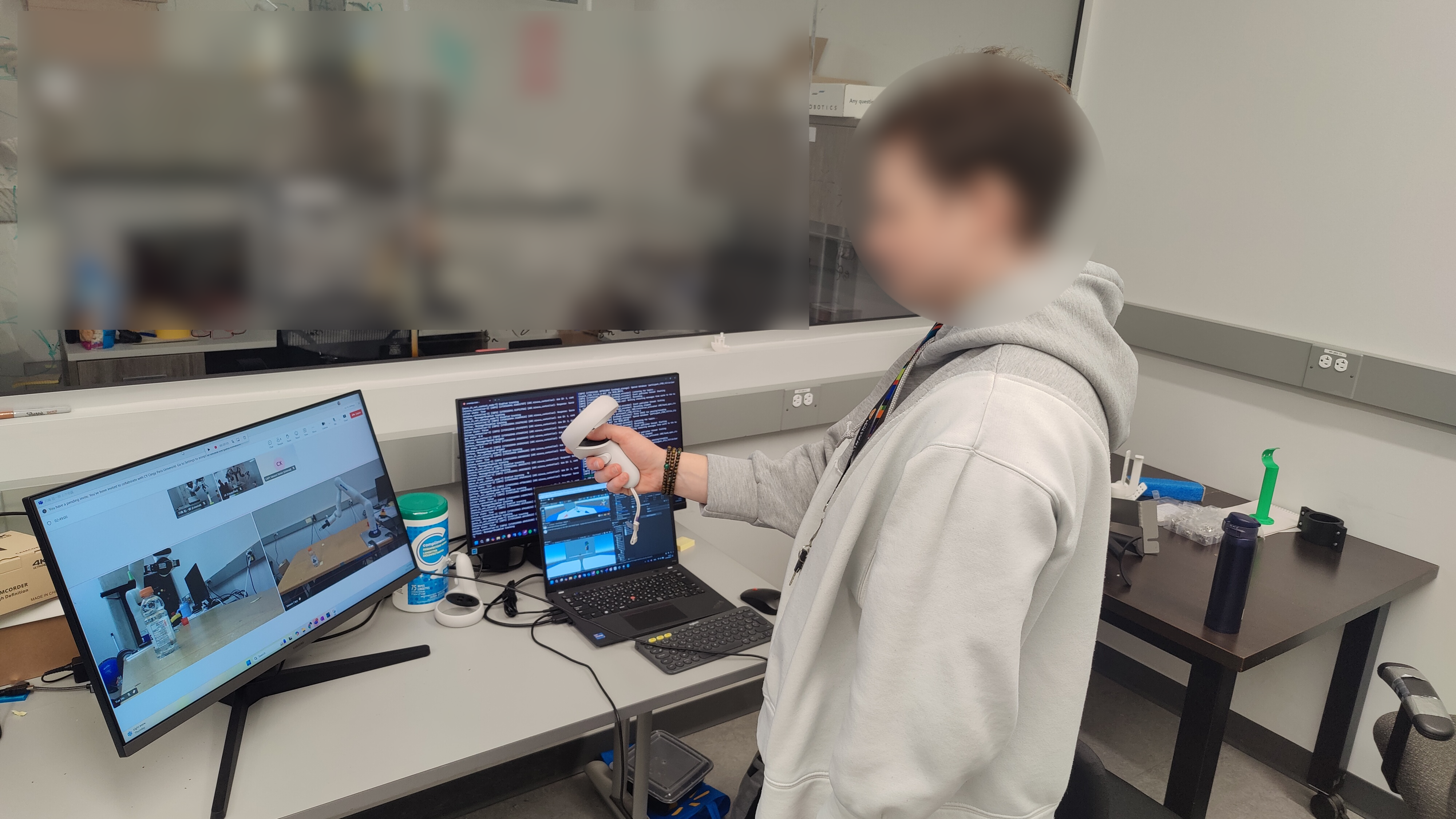}
         \caption{Participant controlling Kinova robot with VR controllers by looking at the monitor.}
         \label{fig:exp-canada-picture}
     \end{subfigure}
     \hfill
     \begin{subfigure}[b]{0.49\textwidth}
         \centering
         \includegraphics[width=\textwidth]{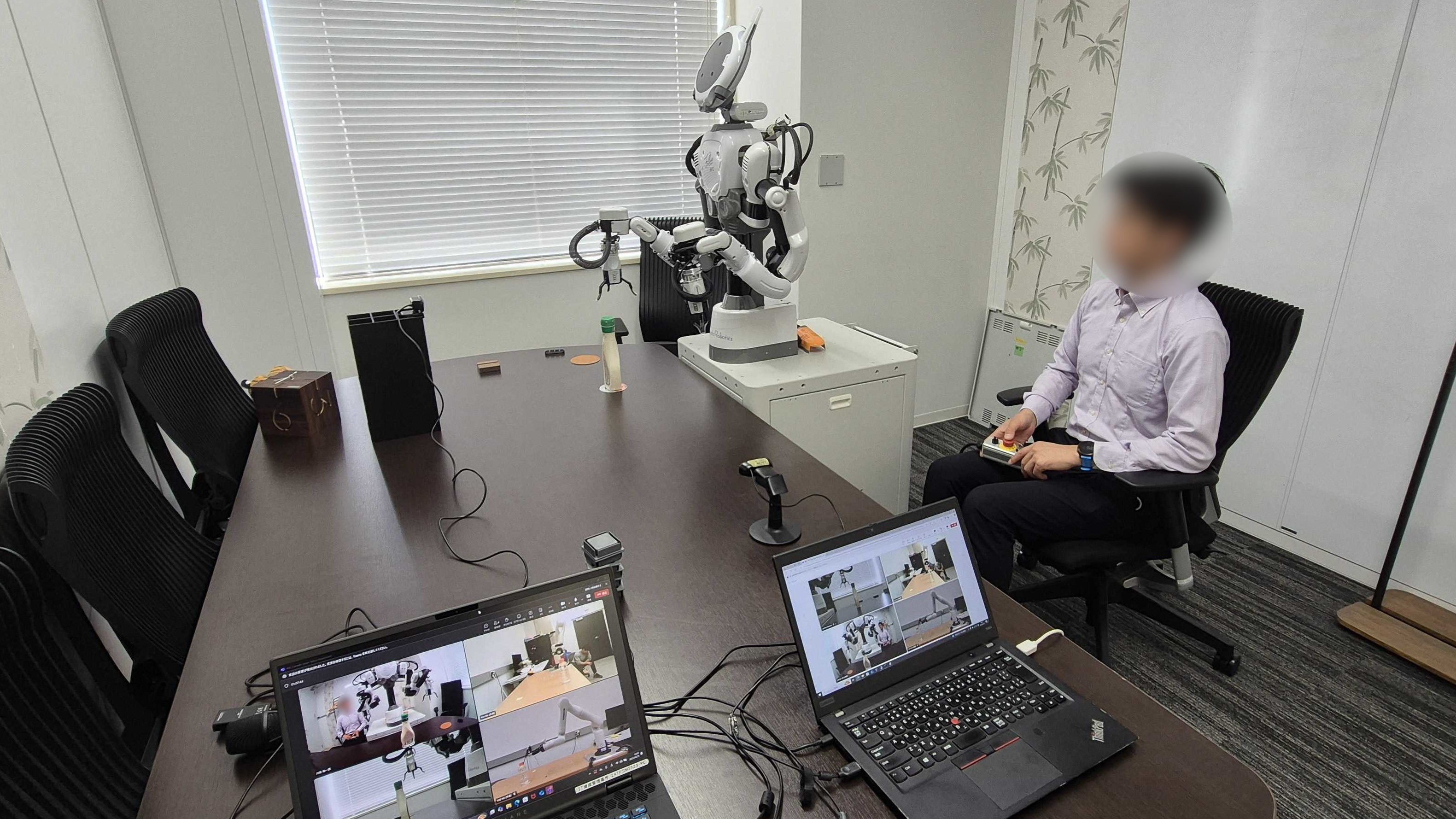}
         \caption{Fillie robot with a researcher positioned to operate the emergency stop button. Each of the two PCs is connected to a webcam.}
         \label{fig:exp-japan-picture}
     \end{subfigure}
        \caption{Overview of the setups used in Canada (left) and Japan (right).}
        \label{fig:exp-pictures}
\end{figure*}

The video stream between the robot and the participants was transmitted via Microsoft Teams, providing two viewpoints: one from the front and one from the side. Figure~\ref{fig:teams-screen} illustrates the control screen displaying all views. To ensure intuitive control, the front view of the Kinova robot was mirrored, aligning the robot’s movement with the left-right axis as perceived by the user. In contrast, for Fillie, the control was reversed directly in the software. Participants did not wear a VR headset, but operated the robots using only VR controllers (see Fig.~\ref{fig:exp-canada-picture}). During the experiment, researcher~1 (see Fig.~\ref{fig:rooms}) wore earphones to communicate with the Japanese team.


Finally, a flexible three-minute timer was allocated for task completion. Participants were instructed to cease the task after this period if it was evident they could not complete it. However, task completion was not among the evaluation criteria. For informational purposes, 96\% of participants completed the task using the Kinova robot, and 75\% using Fillie.

\section{Results}
\subsection{Demography}
The experiment involved 26 participants; two were excluded due to technical issues, resulting in a final sample of \textbf{24 participants} (age 18–28, $M = 21.4$, $SD = 3.6$). Most participants were Canadian-born ($n = 14$, 58\%). Among the foreign-born participants, the average length of residence in Canada was 10 years ($SD = 9$, $Mdn = 6$). The sample included 11 men (46\%), 12 women (50\%), and 1 participant (4\%) who identified as non-binary or preferred not to disclose. Handedness was self-reported: 19 participants (79\%) were right-handed, 4 (17\%) were left-handed and one (4\%) was ambidextrous.

The majority of participants were bachelor's students. Students represented 72\% ($n = 16$) of the sample. The remaining participants included one postdoctoral researcher, one engineer, one event facilitator, and three participants (14\%) who preferred not to answer. Regarding education level, the highest diploma obtained was high school for 17 participants (71\%), bachelor’s degree for 4 (17\%), master’s degree for 2 (8\%), and Ph.\,D. for 1 (4\%).

Nearly half of the participants (47\%, $n = 11$) reported having little to no experience with robots. Four participants (17\%) had between 1–3 hours of experience, three (13\%) had between 3–20 hours, and five (22\%) had between 20–100 hours. The majority of participants ($n = 15$) were occasional or infrequent video game players. Regarding virtual reality (VR), only one participant reported using it regularly.

\subsection{Robot's perception}

\begin{figure}
    \centering
    \includegraphics[width=\linewidth]{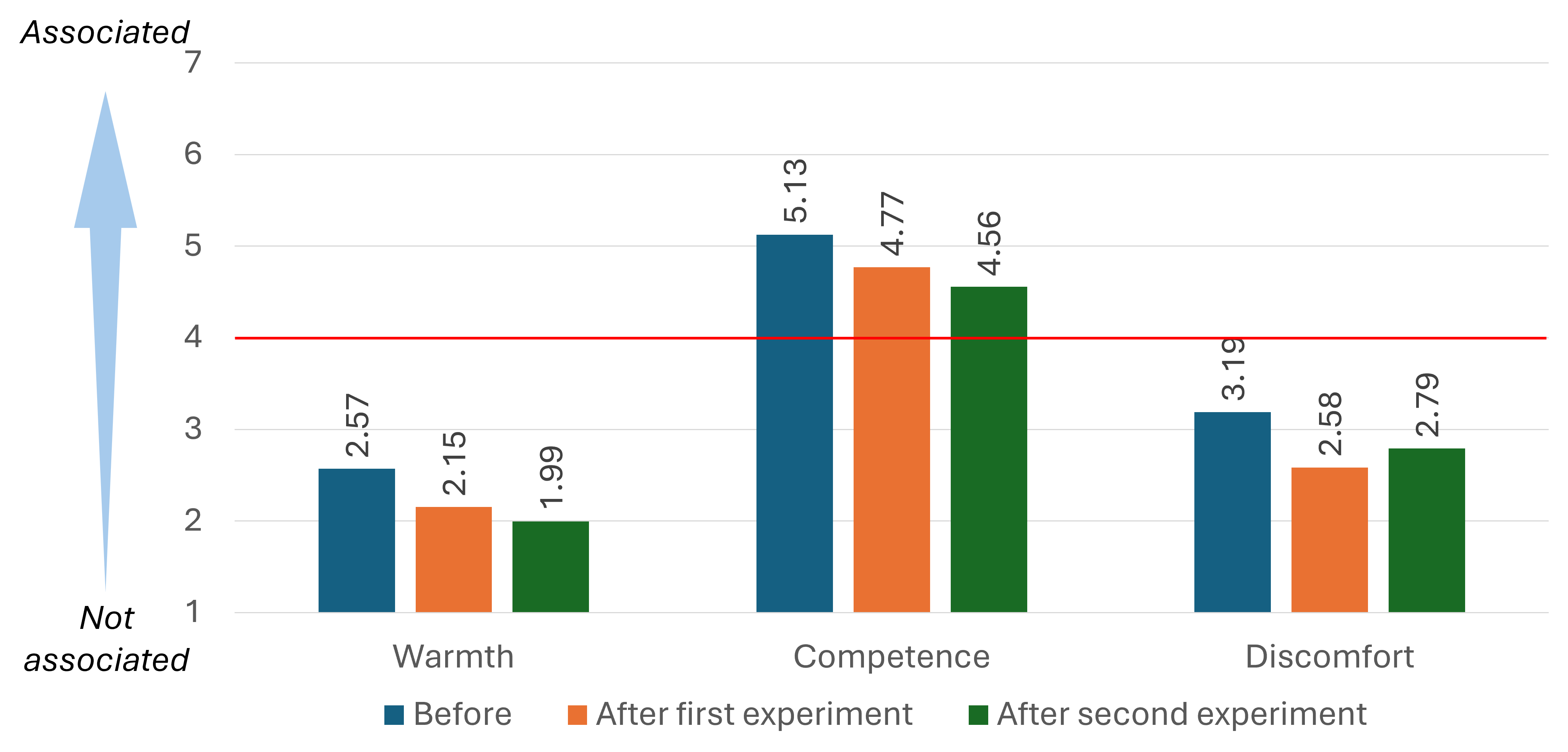}
    \caption{RoSAS (7-Likert scale values) depending on questionnaire. The red line symbolises the mid-value.}
    \label{fig:rosas}
\end{figure}

The RoSAS scale is divided into three dimensions: \textit{warmth}, \textit{competence}, and \textit{discomfort}. Figure~\ref{fig:rosas} illustrates the evolution of these dimensions across the different questionnaires. As the RoSAS results violated the assumption of normality, a non-parametric Friedman test was used. A statistically significant effect of time was found for \textit{warmth} ($\chi^2 = 11.157,~p = .004$) and \textit{discomfort} ($\chi^2 = 12.945,~p = .002$), indicating that participants’ perceptions of the robots evolved throughout the experiment. No statistically significant effect was observed for \textit{competence} ($\chi^2 = 5.255,~p = .072$).

Participants did not perceive the robots as emotionally engaging in terms of \textit{warmth}; instead, they were seen more as mechanical tools. This perception declined after each interaction, likely due to the experimental context in which the robot was used solely for a simple pick-and-place task, without complex or social interaction. This decline was statistically significant between the pre-questionnaire (Q0) and the final one (Q2), as revealed by Holm-corrected post-hoc pairwise comparisons using the Wilcoxon signed-rank test ($W = 55.500,~p = .036$).

Additionally, participants did not report any discomfort related to the robots. On the contrary, \textit{discomfort} levels decreased after interaction, suggesting that participants felt increasingly safe and at ease as the experiment progressed. This finding is consistent with results from the custom questions. The pairwise comparison indicated that the change in \textit{discomfort} was statistically significant between the pre-questionnaire (Q0) and the first one (Q1).

Although no significant change was found in \textit{competence}, scores remained above average, indicating that participants perceived a degree of “intelligence” in the robots, even though the task was relatively simple.

Finally, a Wilcoxon signed-rank test comparing the two robots (for Q1 and Q2) revealed a statistically significant difference only in the \textit{competence} category ($p < 0.05$), with an average score of 4.87 for Kinova and 4.46 for Fillie. However, the effect size was small (Cohen’s $d = 0.411$).

\subsection{Workload}
\begin{figure*}
    \centering
    \includegraphics[width=\linewidth]{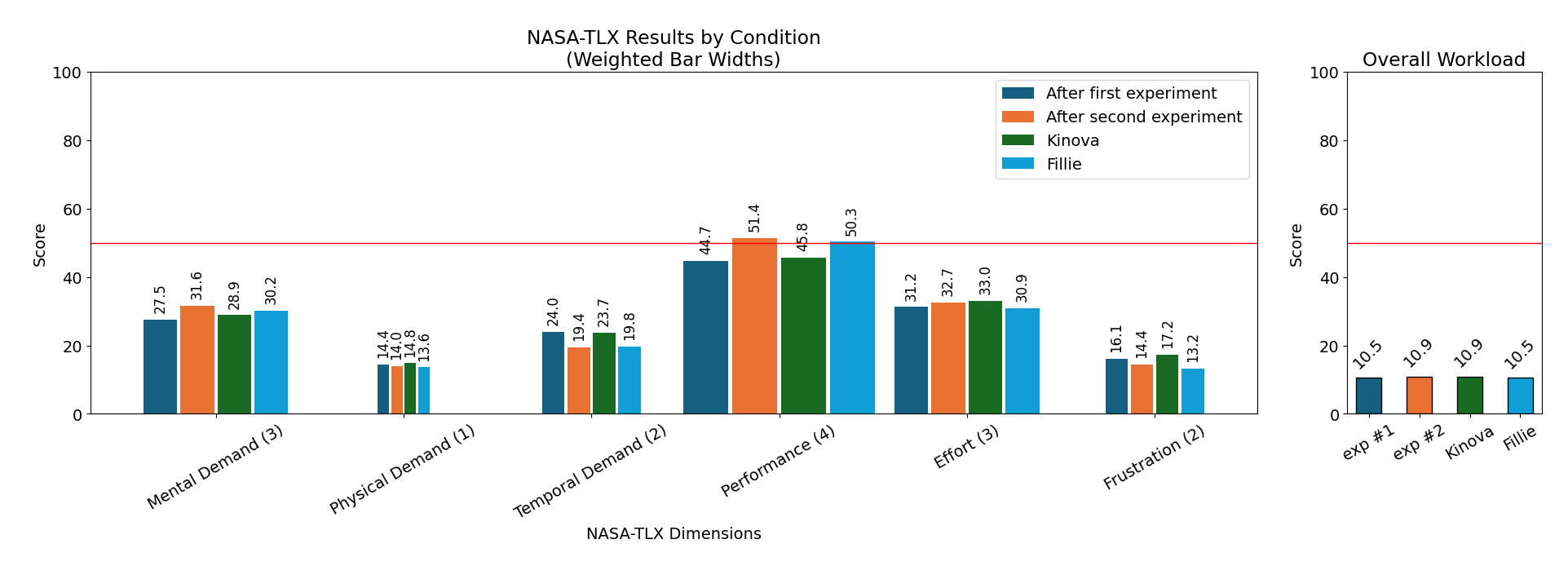}
    \caption{Weighted NASA-TLX score average of all the participants. The width of the bar represents the value of the rounded average weight (indicated in parentheses after each dimension).}
    \label{fig:nasa}
\end{figure*}

Figure~\ref{fig:nasa} presents the score of the NASA-TLX. The overall workload reported for our system was low ($M = 10.7$) and remained consistent over time and across robot conditions. This suggests that the system was not perceived as difficult to use, including in the remote configuration. Participants did not report a high need for concentration while using the system, despite the usability score being below the average SUS benchmark (as discussed later in this section).

A one-way Wilcoxon Signed-Rank Test was conducted, as the NASA-TLX results violated the assumption of normality. The test revealed no significant effects of experiment ID, robot ID, or robot order on any of the NASA-TLX workload dimensions, except for mental demand, where a significant difference was observed between sessions ($W = 33.000$, $p = 0.037$). Mental demand increased by approximately 15\% in the second session.

This may suggest that participants would experience fatigue if required to use the system for extended periods, indicating a potential need for breaks during longer interactions.

\subsection{System's perception}
\begin{figure}
    \centering
    \includegraphics[width=\linewidth]{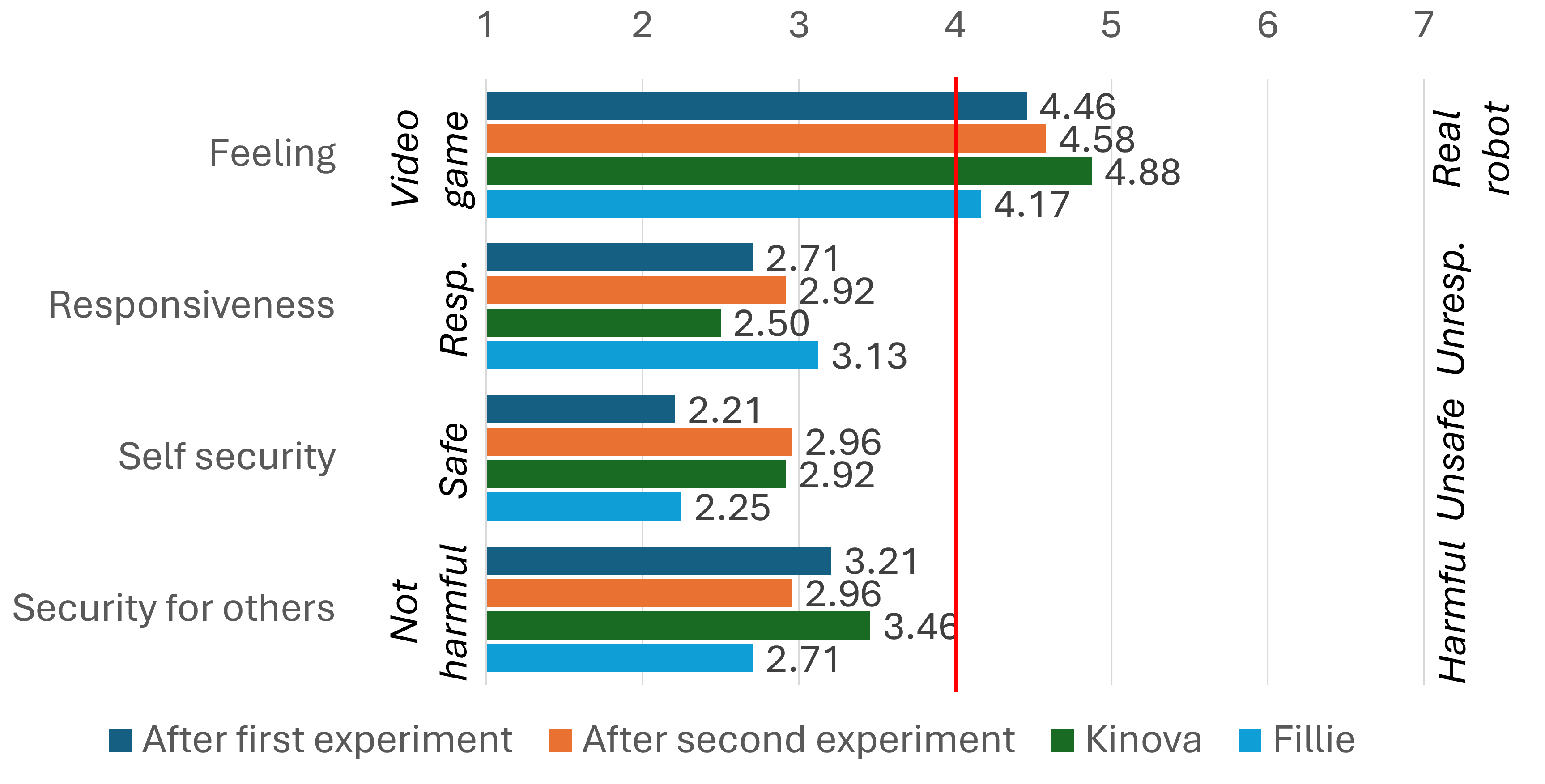}
    \caption{Answer to custom questions (7-Likert Scale values)}
    \label{fig:custom}
\end{figure}

The analysis of the custom questions (see Fig.~\ref{fig:custom}) confirms some intuitive assumptions. However, Wilcoxon Signed-Rank test did not reveal any statistically significant effects, with the exception of the self-safety rating between the first and second experiments. Participants perceived Fillie's control as more similar to a video game than Kinova's (Fillie: 4.17, Kinova: 4.88; where 1 = video game). This may be because Fillie was only visible on a screen, whereas Kinova was physically present in the room, although participants did not directly observe it while controlling it. For both robots, however, the scores remained above the average (4), suggesting that participants found the control to be more robotic than game-like.

Regarding responsiveness, the assumption that the remote robot would feel less responsive is supported by the results (Fillie: 3.13, Kinova: 2.5; where 1 = responsive). The delay introduced by the remote connection likely contributed to this perception. Although the average latency was approximately 1 second, the connection was not stable throughout the session, and delays occasionally increased. Nevertheless, both scores remained below 4, indicating that participants did not feel excessively disturbed by the delay.

As for operator safety near the robot, participants reported feeling safe in both remote conditions (scores below 4; where 7 = harmful). They reported feeling safer with the remote robot (Fillie: 2.71, Kinova: 3.46), possibly due to a reduced sense of physical presence or detachment from the remote environment, however the  Wilcoxon Signed-Rank test did not reveal any statistically significant effects. 

For self-perceived safety, the difference between the first and second experiments was both more pronounced and statistically significant ($W=53.000$, $p=0.048$). While the other differences remained below 10\%, the perceived lack of safety increased by 34\% over time (from 2.21 to 2.96), suggesting a growing concern or awareness after the second experiment. However the safety parameters remain below the average, which means that the participants felt safe, confirming the results of the RoSAS questionnaire. On the other hand, there are no statistically significant effects between the local and remote robots.

Analysis of the optional free-text responses to the first custom question revealed several relevant insights. Two participants expressed difficulty in perceiving depth through the video feed, which negatively impacted their sense of control. One participant noted that controlling the local robot felt \textit{“nothing like a video game for \emph{[them]}”}, in contrast to the remote condition. Another participant described the control of Fillie as feeling real, despite the remote setup: \textit{“It was the concrete knowledge that it was an actual robot that I was controlling that made the experience notable, it instilled a sense of awe”}. Overall, there were no strong preferences between the two robots. 

\subsection{Usability}
The SUS score was 62 after the first experiment and 59 after the second. Depending on the robot used, the SUS score was 63 for the local robot and 58 for the remote one. The average scores across the four conditions are similar and fall within the low marginal range of the SUS scale \cite{sus-brooke-avg,bangor2008empirical}. Considering the average SUS score is 68, our system is reasonably close to the benchmark, especially given that it was developed for research purposes. These results are encouraging for future improvements and development. Moreover, the difference in usability scores between the local and remote conditions is similar to the difference observed between the first and second experiments. This suggests that participants did not perceive the remote operation as less usable, despite the latency introduced by the distance.

\section{Discussion}
In summary, the results revealed no statistically significant differences between the local and remote robot conditions, addressing RQ1 and suggesting that teleoperation can be a viable technical solution, even over long distances. Regarding RQ2, participants did not perceive the robot's operation as burdensome, indicating that teleoperated systems do not necessarily increase workload. However, their long-term use should be further explored in future work.

Even when users are located 10,000 km away, they can still perceive the robot as real, despite Fillie's video game score being close to the average. It is important that users remain aware that the robot’s actions have real-world consequences. This awareness should be taken into account when designing RTo systems. Especially in the healthcare domain, where patients can be involved.

The perception of the environment is also a key factor in enabling operators to perform their tasks smoothly. Participant complaints about difficulty visualising the 3D environment highlight the need to improve the visual interface, making it more natural and immersive. The use of a VR headset is one solution to improve the the perception, usability and workload \cite{whitney2019comparing}.

The current experiment involved only a robotic arm; even in the case of Fillie, the robot was used solely as a single-arm manipulator. A logical next step---aligned with our long-term aim of applying this work in a healthcare context---would be to incorporate a humanoid robot. As the control of such robots is more complex, typically involving coordination of two arms and two legs, it would be valuable to evaluate how users perceive these systems. Fillie represents a suitable intermediate step, as it is semi-humanoid in design. In future applications, the robot could serve as an avatar for the operator to interact directly with patients. In such contexts, a humanoid form would be advantageous in enabling smooth and natural interactions, such as shaking hands, handling objects, or even conveying emotions.

\subsection*{Limitations}
A limitation of our study is the participant sample, which was predominantly composed of bachelor’s students. A more diverse population would allow for the analysis of perceptions across different age groups, potentially yielding more nuanced insights. Another limitation concerns the use of two different robots. However, as the study primarily focused on user perception rather than robot control, this difference is expected to have had only a limited impact on the results. Finally, the simplicity of the task can also be considered a limitation. While it offers a useful first impression of user perception, it may not be sufficient to clearly differentiate between local and remote conditions. A more complex task might require greater concentration and sustained network stability, potentially revealing stronger contrasts in user experience. For example, opening a bottle, building a tower with blocks, or sorting objects. This will be part of our future works.

\section{Conclusion \& Future Works}
\label{conclusion}
In this study, we aimed to evaluate user perception of robot teleoperation over long distances. Our contributions include a dedicated evaluation of this perception in a remote context, supported by a specifically designed protocol. Using a custom-built system and experimental setup, we found no significant differences between the local and remote robot conditions, suggesting that robot teleoperation may serve as a viable alternative to local control.

For future work, we plan to replicate the experiment in the reverse direction (from Japan to Canada) to investigate potential cultural influences on the perception of robot teleoperation between Japanese and Canadian participants. Additionally, we plan to incorporate an intention estimation algorithm to mitigate technical limitations caused by communication delays and user errors. Furthermore, we aim to integrate the video feed inside the VR headset to enable smoother control, using the robot's \textit{point of view}. Finally, we plan to leverage the capabilities of humanoid robots, in particular Fillie, which is semi-anthropomorphic, by enabling control of both arms and the robot's head through the VR interface.

\section*{Acknowledgements}
This work was mainly supported by JST SICORP Grant Number JPMJSC2309, Japan, and partly by the National Research Council (NRC) under Grant NRC-AiP-302-1; we acknowledge the support of Waterloo RoboHub for the use of Kinova Gen3.

\bibliographystyle{IEEEtran}
\bibliography{IEEEabrv,biblio}

\begin{thebibliography}{10}
\providecommand{\url}[1]{#1}
\csname url@samestyle\endcsname
\providecommand{\newblock}{\relax}
\providecommand{\bibinfo}[2]{#2}
\providecommand{\BIBentrySTDinterwordspacing}{\spaceskip=0pt\relax}
\providecommand{\BIBentryALTinterwordstretchfactor}{4}
\providecommand{\BIBentryALTinterwordspacing}{\spaceskip=\fontdimen2\font plus
\BIBentryALTinterwordstretchfactor\fontdimen3\font minus \fontdimen4\font\relax}
\providecommand{\BIBforeignlanguage}[2]{{%
\expandafter\ifx\csname l@#1\endcsname\relax
\typeout{** WARNING: IEEEtran.bst: No hyphenation pattern has been}%
\typeout{** loaded for the language `#1'. Using the pattern for}%
\typeout{** the default language instead.}%
\else
\language=\csname l@#1\endcsname
\fi
#2}}
\providecommand{\BIBdecl}{\relax}
\BIBdecl

\bibitem{teleop97}
A.~Kheddar, C.~Tzafestas, P.~Coiffet, T.~Kotoku, S.~Kawabata, K.~Iwamoto, K.~Tanie, I.~Mazon, C.~Laugier, and R.~Chellali, ``Parallel multi-robots long distance teleoperation,'' in \emph{1997 8th International Conference on Advanced Robotics. Proceedings. ICAR'97}, 1997, pp. 1007--1012.

\bibitem{mohan2021telesurgery-shortsurvey}
A.~Mohan, U.~U. Wara, M.~T.~A. Shaikh, R.~M. Rahman, Z.~A. Zaidi, and M.~T.~A. Shaikh, ``Telesurgery and robotics: an improved and efficient era,'' \emph{Cureus}, vol.~13, no.~3, 2021.

\bibitem{5g}
S.~S. Hussain, S.~M. Yaseen, and K.~Barman, ``An overview of massive mimo system in 5g,'' \emph{IJCTA}, vol.~9, no.~11, pp. 1--12, 2016.

\bibitem{kheddar-ge-jp}
A.~Peer, S.~Hirche, C.~Weber, I.~Krause, M.~Buss, S.~Miossec, P.~Evrard, O.~Stasse, E.~S. Neo, A.~Kheddar, and K.~Yokoi, ``Intercontinental multimodal tele-cooperation using a humanoid robot,'' in \emph{2008 IEEE/RSJ International Conference on Intelligent Robots and Systems}, 2008, pp. 405--411.

\bibitem{oms}
{W}orld~{H}ealth {O}rganization, ``{A}geing and health,'' \url{https://www.who.int/news-room/fact-sheets/detail/ageing-and-health}, 2024, [Accessed 09-04-2025].

\bibitem{paro}
\BIBentryALTinterwordspacing
X.~Wang, J.~Shen, and Q.~Chen, ``How paro can help older people in elderly care facilities: A systematic review of rct,'' \emph{International Journal of Nursing Knowledge}, vol.~33, no.~1, pp. 29--39, 2022. [Online]. Available: \url{https://onlinelibrary.wiley.com/doi/abs/10.1111/2047-3095.12327}
\BIBentrySTDinterwordspacing

\bibitem{Tetsuya-Tanioka2019}
T.~Tanioka, ``Nursing and rehabilitative care of the elderly using humanoid robots,'' \emph{The Journal of Medical Investigation}, vol.~66, no. 1.2, pp. 19--23, 2019.

\bibitem{Martinez-Martin2018}
\BIBentryALTinterwordspacing
E.~Martinez-Martin and A.~P. del Pobil, \emph{Personal Robot Assistants for Elderly Care: An Overview}.\hskip 1em plus 0.5em minus 0.4em\relax Cham: Springer International Publishing, 2018, pp. 77--91. [Online]. Available: \url{https://doi.org/10.1007/978-3-319-62530-0\_5}
\BIBentrySTDinterwordspacing

\bibitem{orylabAvatarRobot-url}
D.~CAFE, ``{A}vatar {R}obot {C}afé,'' \url{https://dawn2021.orylab.com/en/}, [Accessed 09-04-2025].

\bibitem{orylabAvatarRobot-paper}
\BIBentryALTinterwordspacing
Y.~Hatada, G.~Barbareschi, K.~Takeuchi, H.~Kato, K.~Yoshifuji, K.~Minamizawa, and T.~Narumi, ``People with disabilities redefining identity through robotic and virtual avatars: A case study in avatar robot cafe,'' in \emph{Proceedings of the 2024 CHI Conference on Human Factors in Computing Systems}, ser. CHI '24.\hskip 1em plus 0.5em minus 0.4em\relax New York, NY, USA: Association for Computing Machinery, 2024. [Online]. Available: \url{https://doi.org/10.1145/3613904.3642189}
\BIBentrySTDinterwordspacing

\bibitem{t-ro-survey}
K.~Darvish, L.~Penco, J.~Ramos, R.~Cisneros, J.~Pratt, E.~Yoshida, S.~Ivaldi, and D.~Pucci, ``Teleoperation of humanoid robots: A survey,'' \emph{IEEE Transactions on Robotics}, vol.~39, no.~3, pp. 1706--1727, 2023.

\bibitem{tepcoTEPCOApplication}
TEPCO, ``{T}{E}{P}{C}{O} : {A}pplication of {R}obot {T}echnology --- tepco.co.jp,'' \url{https://www.tepco.co.jp/en/decommision/principles/robot/index-e.html}, [Accessed 12-03-2025].

\bibitem{ieeeParisFirefighters}
L.~Peskoe-Yang, ``{P}aris {F}irefighters {U}sed {T}his {R}emote-{C}ontrolled {R}obot to {E}xtinguish the {N}otre {D}ame {B}laze --- spectrum.ieee.org,'' \url{https://spectrum.ieee.org/colossus-the-firefighting-robot-that-helped-save-notre-dame}, 2019, [Accessed 11-03-2025].

\bibitem{surgery}
Y.~Tian, H.~Lv, A.~Jumai, T.~Tuerhong, L.~Zhuang, J.~Huang, J.~Li, P.~Lu, G.~Tao, Y.~Yamauchi, R.~M. Flores, H.~Teng, T.~Chen, and Q.~Luo, ``\BIBforeignlanguage{en}{Ultra-remote robot-assisted right upper lobectomy between the shanghai and kashi prefectures: a case report},'' \emph{\BIBforeignlanguage{en}{J Thorac Dis}}, vol.~16, no.~12, pp. 8823--8830, Dec. 2024.

\bibitem{thomas-wearable}
\BIBentryALTinterwordspacing
T.~M. Kwok, B.~Zhang, and W.~T. Chow, ``A wearable stiffness-rendering haptic device with a honeycomb jamming mechanism for bilateral teleoperation,'' \emph{Machines}, vol.~13, no.~1, 2025. [Online]. Available: \url{https://www.mdpi.com/2075-1702/13/1/27}
\BIBentrySTDinterwordspacing

\bibitem{thomas-gcn}
\BIBentryALTinterwordspacing
T.~M. Kwok, J.~Li, and Y.~Hu, ``Leveraging gcn-based action recognition for teleoperation in daily activity assistance,'' 2025. [Online]. Available: \url{https://arxiv.org/abs/2504.07001}
\BIBentrySTDinterwordspacing

\bibitem{covid19}
\BIBentryALTinterwordspacing
G.~Yang, H.~Lyu, Z.~Zhang, L.~Yang, J.~Deng, S.~You, J.~Du, and H.~Yang, ``Keep healthcare workers safe: Application of teleoperated robot in isolation ward for covid-19 prevention and control,'' \emph{Chinese Journal of Mechanical Engineering}, vol.~33, no.~1, p.~47, Jun 2020. [Online]. Available: \url{https://doi.org/10.1186/s10033-020-00464-0}
\BIBentrySTDinterwordspacing

\bibitem{review-ui-ux}
G.~Sankar, S.~Djamasbi, Z.~Li, J.~Xiao, and N.~Buchler, ``Systematic literature review on the user evaluation of teleoperation interfaces for professional service robots,'' in \emph{HCI in Business, Government and Organizations}, F.~Nah and K.~Siau, Eds.\hskip 1em plus 0.5em minus 0.4em\relax Cham: Springer Nature Switzerland, 2023, pp. 66--85.

\bibitem{nasa-tlx}
\BIBentryALTinterwordspacing
S.~G. Hart and L.~E. Staveland, ``Development of nasa-tlx (task load index): Results of empirical and theoretical research,'' in \emph{Human Mental Workload}, ser. Advances in Psychology, P.~A. Hancock and N.~Meshkati, Eds.\hskip 1em plus 0.5em minus 0.4em\relax North-Holland, 1988, vol.~52, pp. 139--183. [Online]. Available: \url{https://www.sciencedirect.com/science/article/pii/S0166411508623869}
\BIBentrySTDinterwordspacing

\bibitem{sus-brooke}
J.~Brooke, ``Usability evaluation in industry, chap. sus: a “quick and dirty” usability scale,'' 1996.

\bibitem{nars}
\BIBentryALTinterwordspacing
T.~Nomura, T.~Kanda, and T.~Suzuki, ``Experimental investigation into influence of negative attitudes toward robots on human--robot interaction,'' \emph{AI {\&} SOCIETY}, vol.~20, no.~2, pp. 138--150, Mar 2006. [Online]. Available: \url{https://doi.org/10.1007/s00146-005-0012-7}
\BIBentrySTDinterwordspacing

\bibitem{audonnet2024breaking}
F.~P. Audonnet, A.~Hamilton, Y.~Domae, I.~G. Ramirez-Alpizar, and G.~Aragon-Camarasa, ``Breaking down the barriers: Investigating non-expert user experiences in robotic teleoperation in uk and japan,'' \emph{arXiv preprint arXiv:2410.18727}, 2024.

\bibitem{rto-size-height}
\BIBentryALTinterwordspacing
K.~S. Moore, J.~A. Gomer, C.~C. Pagano, and D.~D. Moore, ``Perception of robot passability with direct line of sight and teleoperation,'' \emph{Human Factors}, vol.~51, no.~4, pp. 557--570, 2009, pMID: 19899364. [Online]. Available: \url{https://doi.org/10.1177/0018720809341959}
\BIBentrySTDinterwordspacing

\bibitem{sart}
S.~J. Selcon and R.~Taylor, ``Evaluation of the situational awareness rating technique(sart) as a tool for aircrew systems design,'' \emph{AGARD, Situational Awareness in Aerospace Operations 8 p(SEE N 90-28972 23-53)}, 1990.

\bibitem{ueq}
B.~Laugwitz, T.~Held, and M.~Schrepp, ``Construction and evaluation of a user experience questionnaire,'' in \emph{Symposium of the Austrian HCI and usability engineering group}.\hskip 1em plus 0.5em minus 0.4em\relax Springer, 2008, pp. 63--76.

\bibitem{earth-moon}
J.~Wojtusch, D.~Taubert, T.~Graber, and K.~Nergaard, ``Evaluation of human factors for assessing human-robot interaction in delayed teleoperation,'' in \emph{2018 IEEE International Conference on Systems, Man, and Cybernetics (SMC)}, 2018, pp. 3787--3792.

\bibitem{ros2}
\BIBentryALTinterwordspacing
S.~Macenski, T.~Foote, B.~Gerkey, C.~Lalancette, and W.~Woodall, ``Robot operating system 2: Design, architecture, and uses in the wild,'' \emph{Science Robotics}, vol.~7, no.~66, p. eabm6074, 2022. [Online]. Available: \url{https://www.science.org/doi/abs/10.1126/scirobotics.abm6074}
\BIBentrySTDinterwordspacing

\bibitem{zmq}
``{Z}ero{M}{Q} --- zeromq.org,'' \url{https://zeromq.org}, [Accessed 10-03-2025].

\bibitem{capy2022expanding}
S.~Capy, L.~Rincon, E.~Coronado, S.~Hagane, S.~Yamaguchi, V.~Leve, Y.~Kawasumi, Y.~Kudou, and G.~Venture, ``Expanding the frontiers of industrial robots beyond factories: Design and in the wild validation,'' \emph{Machines}, vol.~10, no.~12, p. 1179, 2022.

\bibitem{rosas}
\BIBentryALTinterwordspacing
C.~M. Carpinella, A.~B. Wyman, M.~A. Perez, and S.~J. Stroessner, ``The robotic social attributes scale (rosas): Development and validation,'' in \emph{Proceedings of the 2017 ACM/IEEE International Conference on Human-Robot Interaction}, ser. HRI '17.\hskip 1em plus 0.5em minus 0.4em\relax New York, NY, USA: Association for Computing Machinery, 2017, p. 254–262. [Online]. Available: \url{https://doi.org/10.1145/2909824.3020208}
\BIBentrySTDinterwordspacing

\bibitem{nasa-tlx-computation}
S.~G. Hart, ``Nasa task load index (tlx): Paper and pencil package-volume 1.0,'' 1986.

\bibitem{sus-brooke-avg}
J.~Brooke, ``Sus: a retrospective.'' \emph{Journal of usability studies}, vol.~8, no.~2, 2013.

\bibitem{bangor2008empirical}
A.~Bangor, P.~T. Kortum, and J.~T. Miller, ``An empirical evaluation of the system usability scale,'' \emph{Intl. Journal of Human--Computer Interaction}, vol.~24, no.~6, pp. 574--594, 2008.

\bibitem{whitney2019comparing}
D.~Whitney, E.~Rosen, E.~Phillips, G.~Konidaris, and S.~Tellex, ``Comparing robot grasping teleoperation across desktop and virtual reality with ros reality,'' in \emph{Robotics research: the 18th international symposium ISRR}.\hskip 1em plus 0.5em minus 0.4em\relax Springer, 2019, pp. 335--350.

\end{thebibliography}

\vspace{12pt}

\end{document}